# Construction Site Scaffolding Completeness Detection Based on Mask R-CNN and Hough Transform

Pei-Hsin Lin, Jacob J. Lin, Shang-Hsien Hsieh
National Taiwan University, Taiwan
peihsin@caece.net

**Abstract.** Construction site scaffolding is essential for many building projects, and ensuring its safety is crucial to prevent accidents. The safety inspector must check the scaffolding's completeness and integrity, where most violations occur. The inspection process includes ensuring all the components are in the right place since workers often compromise safety for convenience and disassemble parts such as cross braces. This paper proposes a deep learning-based approach to detect the scaffolding and its cross braces using computer vision. A scaffold image dataset with annotated labels is used to train a convolutional neural network (CNN) model. With the proposed approach, we can automatically detect the completeness of cross braces from images taken at construction sites, without the need for manual inspection, saving a significant amount of time and labor costs. This non-invasive and efficient solution for detecting scaffolding completeness can help improve safety in construction sites.

## 1. Introduction

In recent years, construction site safety has been a topic of increasing concern in Taiwan. The construction industry is regarded as high risk because construction workers are often exposed to dangerous working conditions, such as working at heights and in confined spaces. The risk of accidents in the construction industry is much higher than in other industries, which makes it important to improve safety standards and develop effective safety monitoring methods. Taking Taiwan's construction industry as an example, statistics from the past decade indicate that there were 3,157 worker deaths in industrial safety accidents, nearly half of which were due to falls from heights. These numbers are alarming and highlight the urgent need to improve safety management on construction sites.

Despite the many safety regulations in place, such as the need to wear safety belts when working at heights and the requirement for protruding steel bars to be equipped with protective covers, many workers still do not comply with them properly. According to the reports of Taiwan's Occupational Safety and Health Administration (https://www.osha.gov.tw/48783/), many accidents occur due to failure to comply with safety standards, and this negligence will cause irreparable results. In the case of scaffolding, cross braces are often disassembled for convenience but forgotten to be reinstalled, which may cause workers to fall from the gaps in the scaffolding when working at heights.

Developing deep learning and computer vision technology has provided new opportunities for construction site safety monitoring. The use of computer vision technologies and deep learning in construction site safety monitoring has the potential to improve safety standards by detecting safety hazards in real-time and preventing accidents before they happen. With the use of computer vision technology, it is possible to monitor construction sites remotely, which can help to reduce the risk of accidents and save lives.

The purpose of this research is to develop a real-time safety monitoring method using deep learning and computer vision technology to detect incomplete scaffolding structures in construction sites, and subsequently alert workers and supervisors to potential safety hazards. This method employs a combination of computer vision techniques and deep learning algorithms to ensure the integrity of scaffolding.

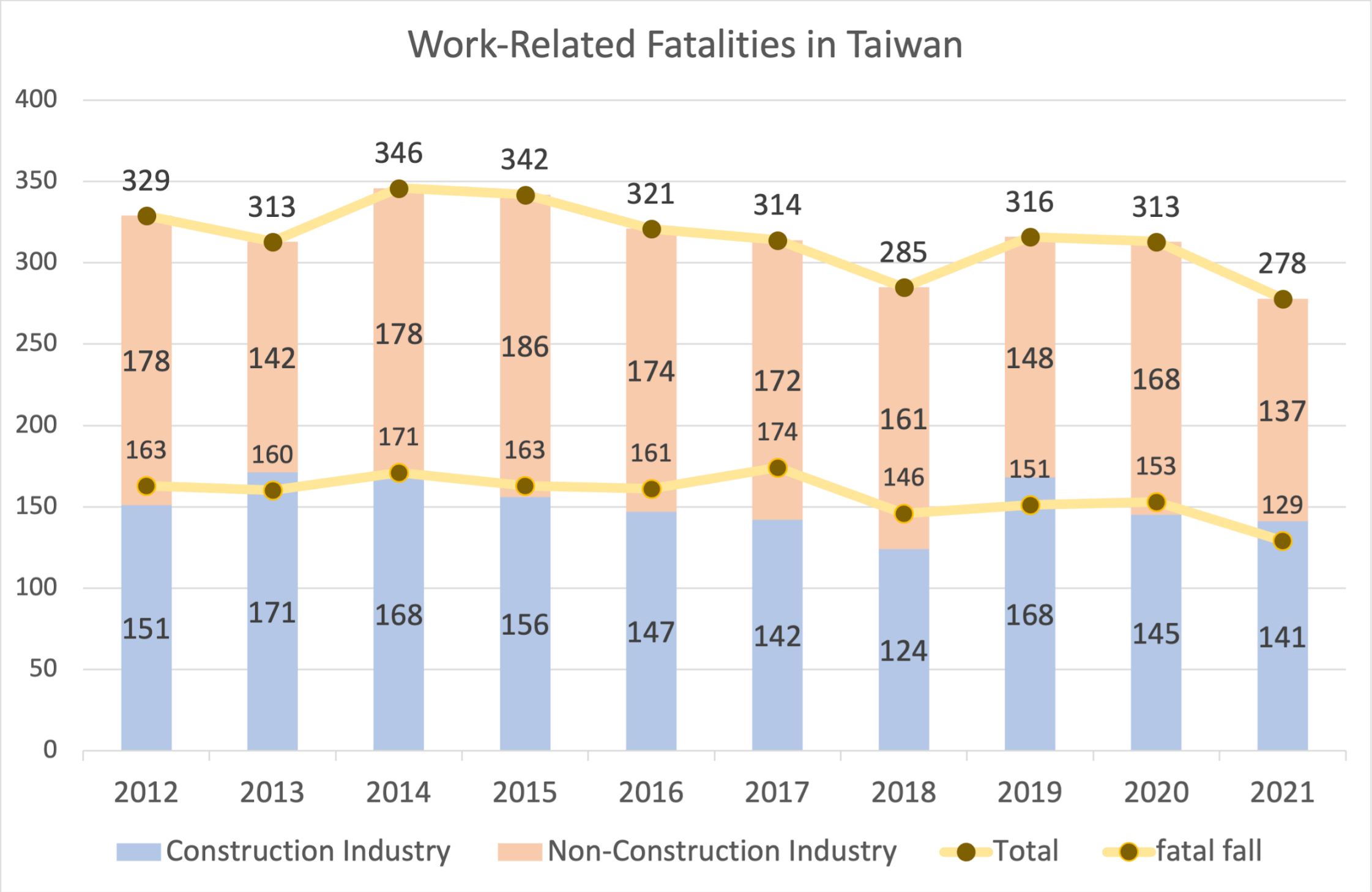

Figure 1: occupational accident fatality rates and causes from 2013 to 2021.

## 2. Related Research

In recent years, there have been numerous research papers discussing construction site safety issues, and many of them have adopted research methods such as deep learning, computer vision, Building Information Modeling (BIM), or point cloud technology. Khan et al. (2022) have developed a real-time monitoring method for falling from heights that utilizes computer vision and Internet-of-Things (IoT) to detect whether workers are wearing personal protective equipment. We all hope to reduce the workload of site managers through real-time alert systems, but what sets our study apart is that it aims to ensure worker safety by monitoring whether cross braces have been dismantled.

When it comes to detecting cross braces, the most difficult part is that cross braces are too thin and lack significant features, making it challenging to achieve accurate detection through image segmentation. Li (2019) proposed a method for segmenting cables based on their color and width, which can achieve a mean precision of approximately 50%, which may help segment cross brace. Wang (2019) introduced an inspection method for scaffold work platforms that uses 3D point cloud data to first locate vertical scaffold components, and then detect horizontal work platforms based on the histogram of the Z values of the point cloud data. This approach can further inspect items such as toe boards and guardrails and overcome the limitations of manual inspection. The recognition method and results are admirable, but unfortunately, point cloud-based recognition is difficult to apply to real-time warning systems on construction sites because creating point clouds requires much manpower and time.

Fang et al. (2019) developed a computer-vision approach to address the problem of individuals traversing structural supports during construction activities, which can lead to an increased risk of falls and injuries. The approach uses a Mask Region-Based Convolutional Neural Network to detect individuals on structural supports and recognize their relationship with concrete and steel supports. This proposed method was validated using a database of photographs and achieved high recall and precision rates. The results indicate that the developed computer-vision approach can be used to identify unsafe behavior and provide feedback to prevent its recurrence.

Kolar et al. (2018) mentioned the construction industry's long-standing concern for safety. Currently, safety inspections rely heavily on human efforts, which are limited on-site. To improve safety performance, this paper proposes using information and sensing technologies for automated and intelligent monitoring and inspection to supplement current manual inspection practices. They focus on the high incidence of falls from heights in construction sites and develop a safety guardrail detection model based on convolutional neural networks(CNN). The proposed CNN-based guardrail detection model achieved high accuracy of 96.5% with augmented data sets generated through image synthesis and transfer learning to extract basic features for neural network training. This approach has promising applications in construction job site safety monitoring.

In summary, to address the issue of falling from heights, various methods have been proposed by different individuals. Some suggest detecting personal protective equipment as a solution, while others utilize computer vision technology to monitor unsafe behavior on construction sites. Our study posits that the integrity of scaffolding components can indirectly or directly contribute to the occurrence of fall accidents. Therefore, it focuses on inspecting the integrity of cross braces as a preventive measure against falls and proposes a warning system to raise real-time awareness among on-site personnel about potential hazards.

## 3. Methodology

The first step in identifying a cross brace is to analyze its shape. In a 2D image, it is a shape composed of two intersecting straight lines. This shape can be challenging to recognize in image recognition because there are many similar and thin objects to detect, but the benefit is that the lines and intersections can be used as features for identification. Since construction site images often have cluttered backgrounds, image segmentation is used to isolate the area that needs to be recognized first. The next step is to use the Hough transform, which is a feature extraction technique that can recognize specific shapes in an image, such as lines, circles, and more. In this case, we use it to detect lines and extract them from the specified region, which includes the cross brace and uprights of the scaffold as they are composed of lines in 2D images. Finally, we use k-means clustering to separate the lines obtained in the previous step into two clusters and mark the intersection points, representing the cross brace's location.

### 3.1. Geometry of Scaffold

A scaffold is composed of many components, such as work platforms, uprights, and cross braces, as shown in Figure 3. If the scaffold is broken down into units, it has a width of 762 millimeters (net width of 719.3 millimeters) with two 30-centimeter wide work platforms placed in the middle. The gap between the two work platforms is within 3 centimeters, or a work platform wider than 50 centimeters can be placed in the middle to ensure that the net width between the work platforms and the uprights of the scaffold is below 20 centimeters, leaving no possibility for falling.

According to regulations in Taiwan, cross braces must be installed if the scaffold is over 2 meters high. If the scaffold needs to be dismantled for operational needs (within the scope of one floor's height), it should be designed by professionals with expertise and experience to ensure the safety of the operation after dismantling. During the dismantling process, safety measures such as safety nets should be used to maintain the safety of working at heights. After the operational needs are completed, the dismantled cross braces should be restored.

With this structure, we can proceed with the segmentation and line detection process unit by unit.

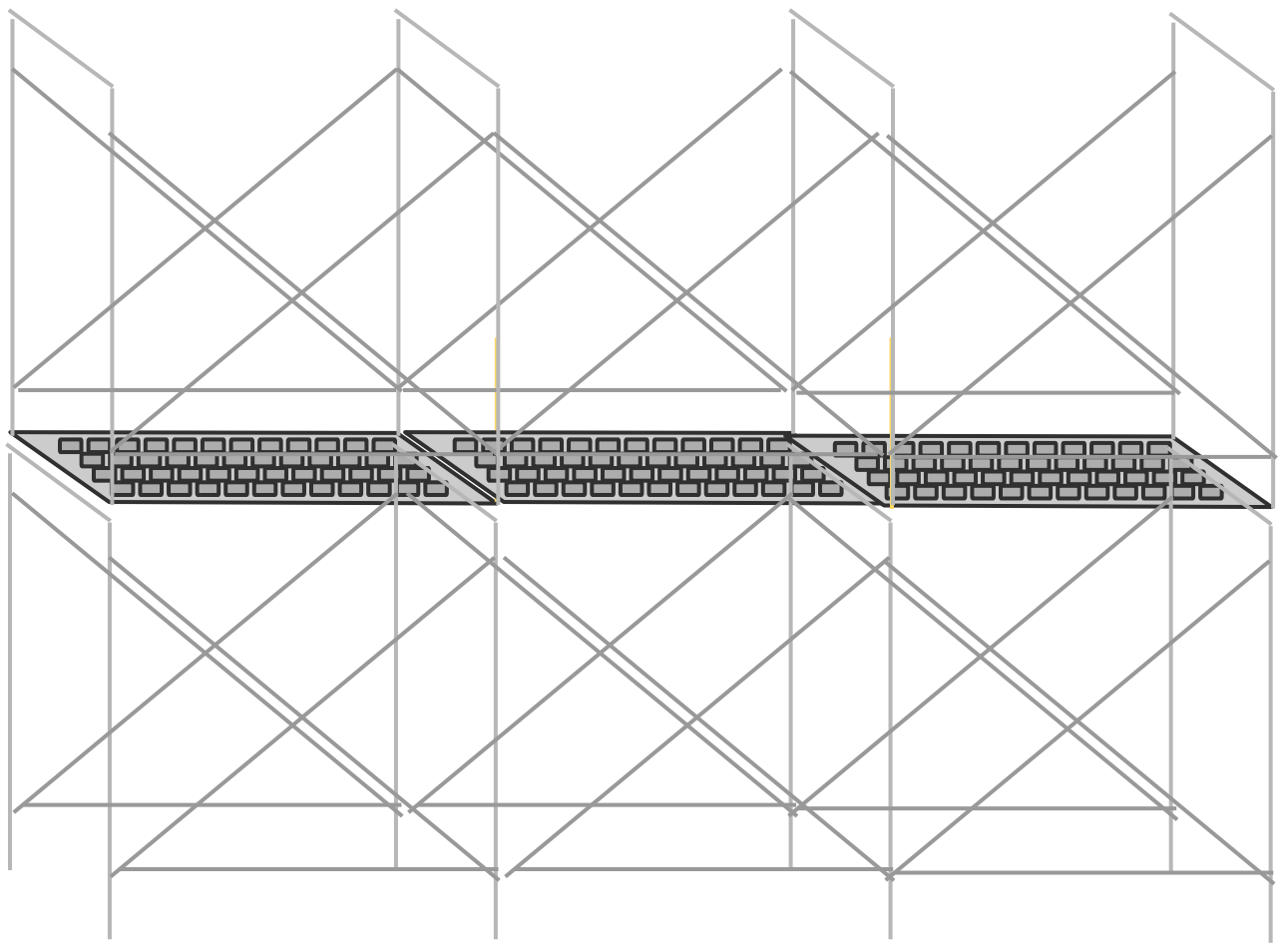

Figure 2: Geometry of Scaffold

### 3.2. Dataset Preparation

In the dataset, 372 photos of scaffolding were split into training and testing data in an 8:2 ratio. The Computer Vision Annotation Tool (CVAT) was used for annotating the images, and the annotated results were exported in the format of the coco dataset as a json file. The reason for annotating by the unit is better for analyzing the components of scaffolding compared to annotating larger areas.

An attempt was made to annotate only cross braces and train a model specifically for them, but it was difficult to detect, with bounding box results of only 41.35% and segmentation results of only 1.85%. To improve this, a deep prediction model was also included, forming a model with 4 channels (RGB-D). However, the results were still very unsatisfactory, with a segmentation average precision (AP) of only 7.85%. Such precision would result in many errors and uncertainties in practice, so this approach was ultimately not used.

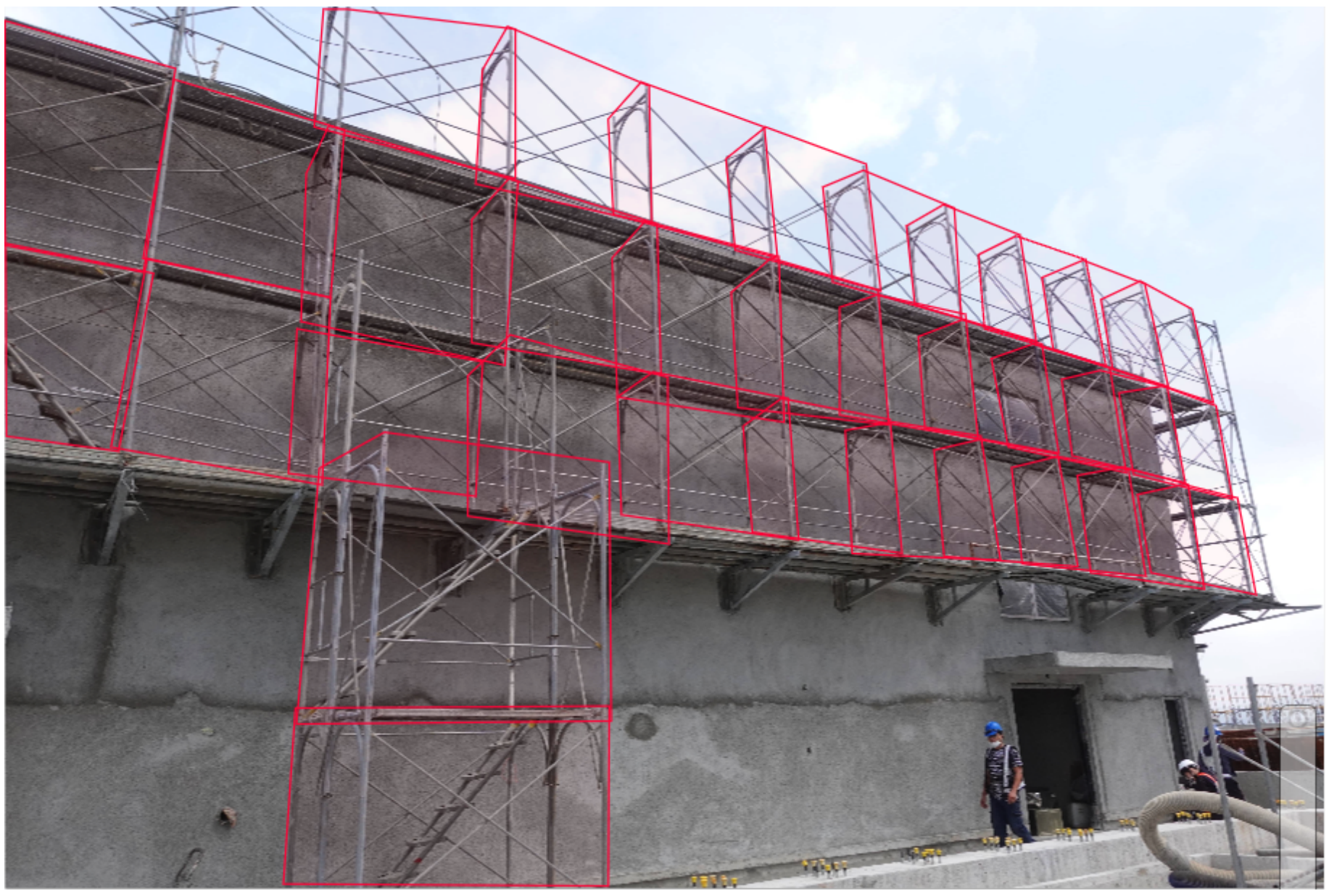

Figure 3: Sample annotations

### 3.3. Image Segmentation

At this stage, the main objective is to segment the area to be recognized, which also makes it easier for the Hough transform to be performed later. Detectron2 is used for image segmentation. Detectron2 is an open-source object detection library developed by Facebook AI Research (FAIR), built on top of the PyTorch deep learning framework. It provides a modular and flexible framework for building and training state-of-the-art object detection models. Detectron2 is designed with a modular architecture, which allows users to easily customize and extend the library for their specific needs. This modularity also makes it easier to integrate new models, datasets, and algorithms into the library. One of the key strengths of Detectron2 is its support for a wide range of object detection models, including Faster R-CNN, RetinaNet, and Mask R-CNN. These models are pre-configured and can be easily fine-tuned or trained from scratch using the library's APIs. Detectron2 also includes several useful features and tools, such as data augmentation, distributed training, and support for multiple GPUs. Detectron2 includes a powerful visualization toolkit, which allows users to easily visualize the results of their models and analyze their performance. Overall, Detectron2 is a powerful and flexible library for building and training object detection models, with a strong focus on modularity, flexibility, and ease of use.

### 3.4. Hough Transform

Hough Transform is a popular computer vision technique used for detecting geometric shapes in images, particularly lines, and circles. It was proposed by Paul Hough in 1962 as a method for detecting particle tracks in bubble chamber photographs. The basic idea of the Hough Transform is to represent shapes in an image as points in a parameter space, and then to detect them by finding the clusters of points in the parameter space.

The Hough Transform algorithm begins by analyzing each point in the input image to determine if it belongs to a shape. For example, in line detection, each edge pixel in the image is examined to determine if it lies on a line. If the point belongs to a shape, then it is used to vote for the shape's parameters in the parameter space. The parameters depend on the type of shape being detected. For example, for line detection, the parameters are the slope and y-intercept of the line. After all the points have been voted, the algorithm looks for clusters of votes in the parameter space. The clusters represent the shapes present in the image. The location and size of the clusters correspond to the site and orientation of the shapes in the image.

Hough Transform has several advantages over other shape detection methods. It is robust to noise and can handle shapes with gaps and other imperfections. It is also computationally efficient, as the algorithm only needs to look for clusters of votes in the parameter space, rather than searching the entire image for shapes.

However, Hough Transform also has some limitations. It can only detect shapes that can be represented parametrically, such as lines and circles. It is also sensitive to the choice of parameters used to represent the shapes and may require tuning for different types of images.

In summary, Hough Transform is a powerful technique for detecting geometric shapes in images. Its ability to handle noise and imperfections, and its computational efficiency, make it a popular choice for many computer vision applications.

### 3.5. Calculate intersections

Before the intersections can be found, the lines obtained by Hough Transform need to be separated into two groups. After the angles of all line segments are extracted, the angles are transformed to be within [0, pi] radians, and then converted into coordinates on the unit circle. Then, the k-means algorithm is used to divide these points into 2 clusters, and the line segments are segmented based on the assigned labels. Finally, two groups of segmented line segments can be obtained. Next, the lines from different groups are paired in a loop, and the intersections are calculated.

Once the intersection points have been found, we can start drawing the lines on the original image. The part where the lines are drawn uses a for loop to iterate over the collection of line segments called 'lines'. Each line segment first extracts the polar coordinates (rho, theta) from its representation, and then uses trigonometric functions to calculate the Cartesian coordinates (x1, y1) and (x2, y2) in the rectangular coordinate system.

Specifically, for each line segment, the function that calculates the slope (a) and intercept (b) of the line based on its polar coordinates (rho, theta) uses the equation of the line to calculate two points (x0, y0) and (x1, y1) on the line. It then calculates another point (x2, y2) on the line at a fixed length of 1000, using the two points (x0, y0) and (x1, y1). Finally, the cv2.line function is used to draw the line on the image.

## 4. Result and Discussion

In the image segmentation part, the average precision of the bounding box reached 76.376%, while the segmentation achieved 74.189%. After multiple fine-tuning iterations, the best parameters were obtained with the 'configs' using 'mask_rcnn_R_101_FPN_3x.yaml'. In addition, the learning rate and batch size were set to 0.0001 and 4, respectively. Also, data augmentation is applied to it.

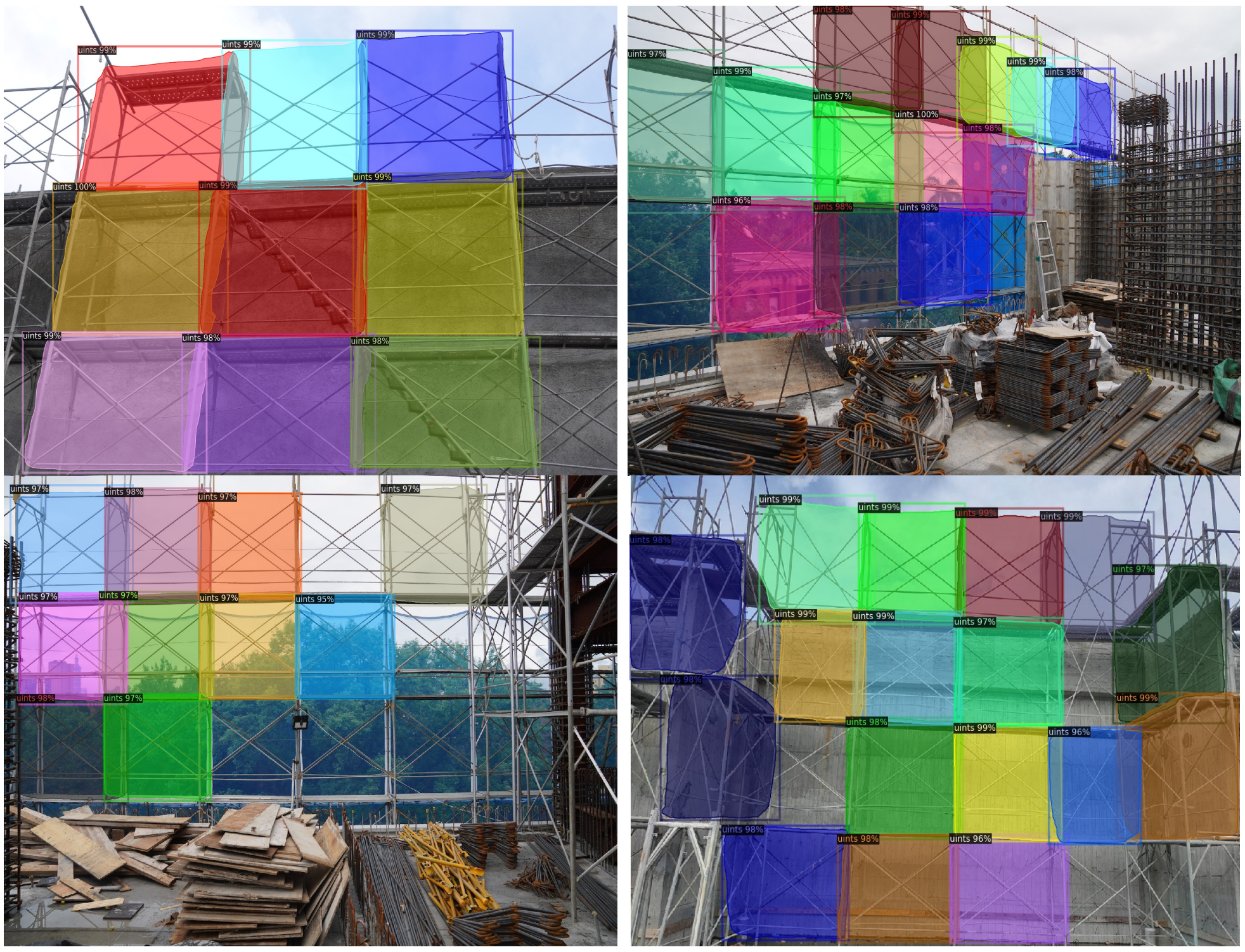

Figure 4: the result of image segmentation

| | AP50 | Average Recall | F1 score |
|---|---|---|---|
| Bounding Box | 76.376% | 64% | 0.6964 |
| Segmentation | 74.189% | 63.2% | 0.6825 |

Table 1: Quantitative indicator of image segmentation

After the prediction is completed, the predicted instances can be outputted. Here, we output one photo per instance (i.e. per unit) to ensure that each instance can be recognized independently. The outputted photos have black pixels in all areas except the mask, so if there are multiple instances in one photo, multiple results will be outputted. Before moving on to the next step, it is necessary to use methods such as cv2.adaptiveThreshold to convert the photo into black and white form to reduce noise and segment the foreground.

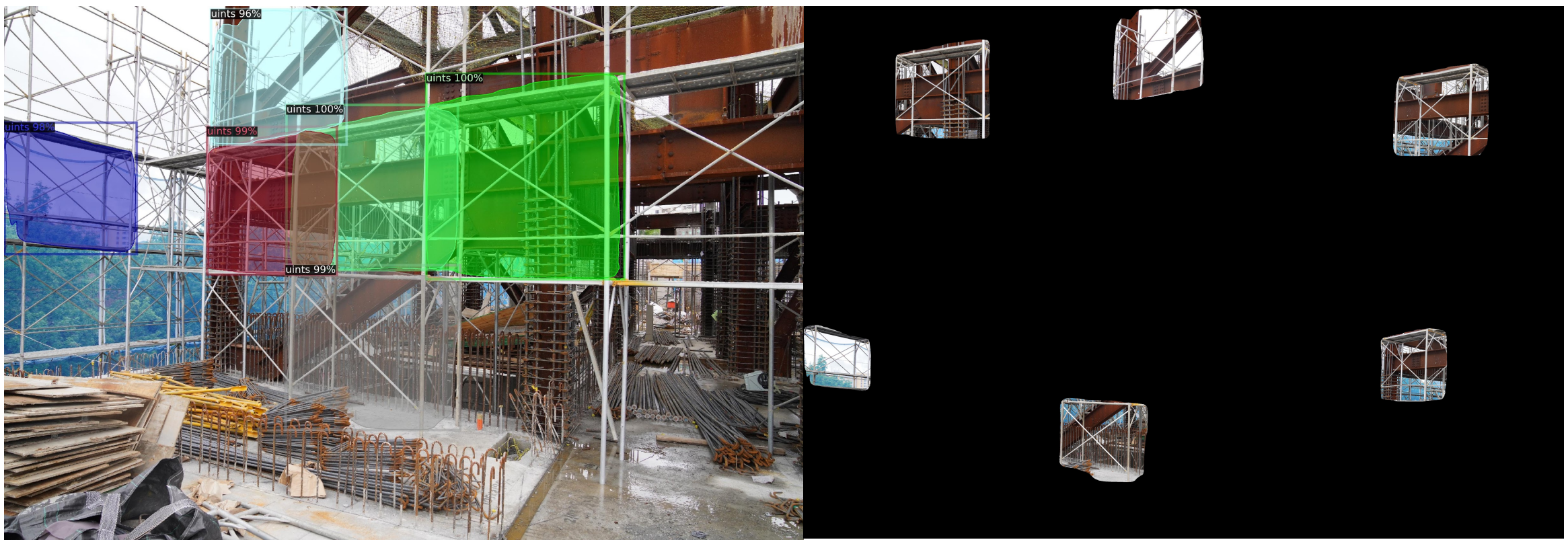

Figure 5: There are 6 instances in this image, so six photos will be outputted.

By applying the Hough transform to these outputs, we can calculate the intersection points to obtain the final results. For the Hough transform part, some parameters need to be adjusted. For example, adjusting the threshold parameter can affect the number of detected lines in the final result. However, in this case, the parameters do not need to be adjusted too much to obtain the result. After confirmation, 129 instances were detected from 29 testing data, and almost 85% of cross braces were detected. Of course, some scaffold units may have been missed due to the image segmentation part having a precision of about 75%. After completing these predictions, the detected cross brace points' disappearance represents the cross braces' disassembly.

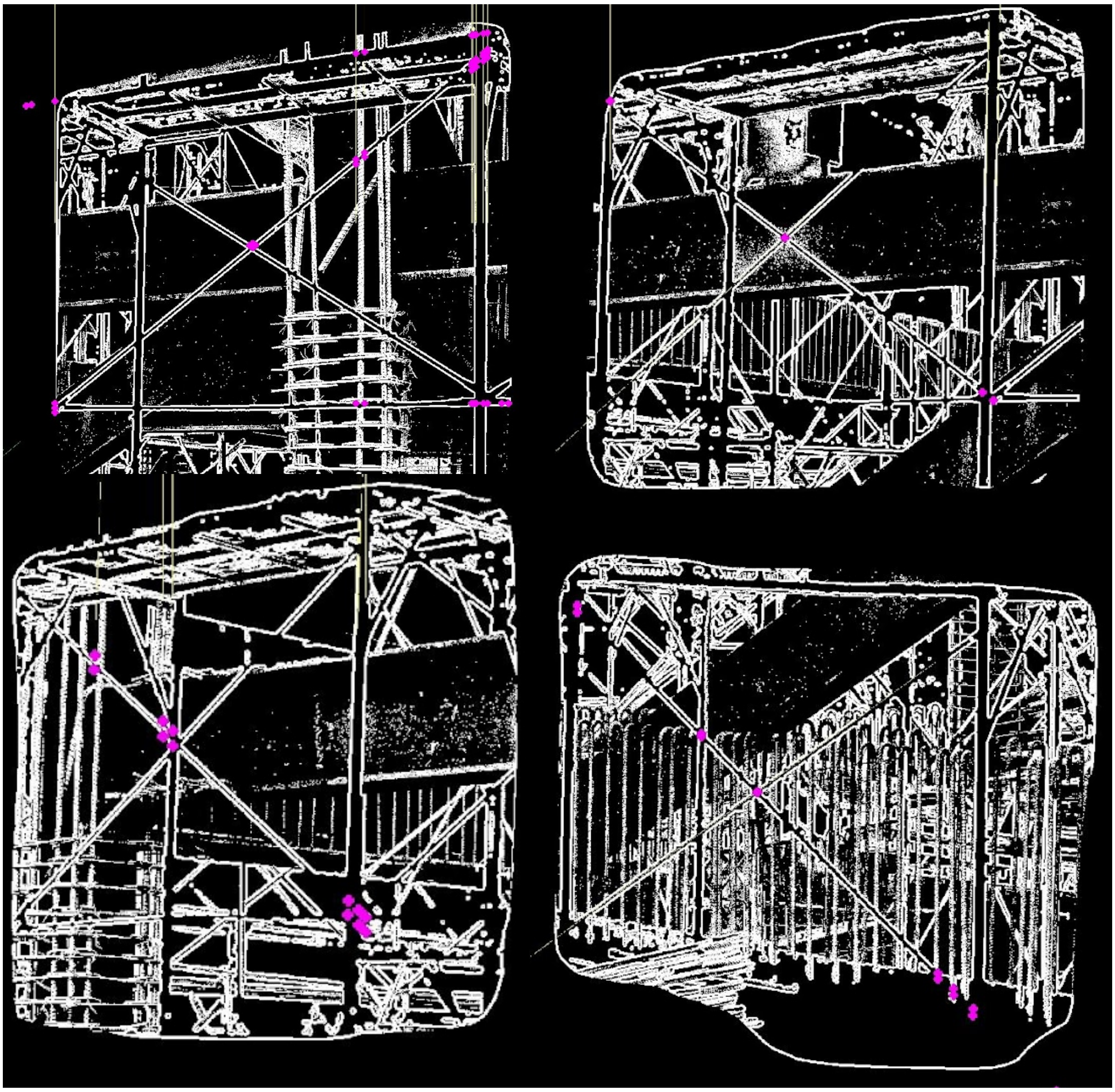

Figure 6: Detection results

We have identified several future directions to improve the current result.

1. Dataset expansion and augmentation: The current average precision is around 75%, and it can be improved by adding more data (currently only 372 images) and using methods such as Generative Adversarial Networks (GANs) to generate synthetic data that closely resembles real-world images. By introducing these synthetic images, the model can improve its ability to generalize and perform well on unseen data. Similar approaches can enhance data completeness and improve the average precision, resulting in higher reliability in subsequent detections.

2. Real-world scenario evaluation: It is still unclear whether this method can run smoothly on CCTV in construction sites. After all, there are many ways to dismantle cross braces, and false alarms may cause trouble for on-site managers. In the future, periodic detections can be conducted, and if there is a lack of intersection points compared to the previous image, an alarm can be triggered.

3. Integration with depth feature: Currently, the use of the Hough transform cannot uniquely identify the intersection point of the cross brace in the entire image, and there may be many other intersection points. The intersection points of uprights and cross braces can be removed using k-means, but there are still other non-essential results due to the cluttered background of construction site photos. Knowing the sizes of the components of the scaffolding unit, if a depth prediction can be introduced to distinguish the scene depth differences between cross braces and other non-essential results, it may improve the results.

## 5. Conclusion

Construction sites are known to be hazardous environments with many unsafe practices or conditions that can cause accidents or fatalities. Currently, most construction sites rely on manual inspections by safety personnel. However, this approach is prone to human error and can be time-consuming and labor-intensive. Additionally, there are numerous incidents of falling accidents every year, with various causes. Dismantling of cross braces is one of the factors that can lead to falling accidents. The primary objective of this study is to leverage deep learning and computer vision to enhance construction site safety management, reduce the burden on safety inspectors, and prevent accidents.

In practical applications, real-time monitoring is achieved using closed-circuit television (CCTV) cameras that capture images of the construction site at regular intervals. These images are then subjected to image segmentation, which helps to segment the scaffolding into image units with an average precision of around 75%. In addition, analyzing the geometric shape of the scaffolding suggests that line detection is a more effective method than cross braces segmentation for locating the position of cross braces. This is because cross braces are usually straight lines that intersect at right angles, making them easily detectable using line detection techniques. By utilizing line detection methods, such as the Hough transform, the accuracy of cross braces detection can be further improved. At this point, k-means clustering is applied to the detected lines to further calculate intersection points, enabling us to locate the position of cross braces through these identified intersection points. The accuracy of cross braces detection by this approach is approximately 85%.

The ultimate aim of this research is to provide accurate warnings and alerts to site personnel in the event of safety hazards, allowing them to take immediate actions. The alerts could be in the form of audible or visual signals that notify workers to avoid certain areas or take safety precautions. The combination of deep learning and computer vision provides a powerful tool to help prevent accidents on construction sites. By automating visual safety inspections, safety personnel can focus their efforts on more complex safety issues.

In conclusion, this study highlights the potential of deep learning and computer vision in enhancing safety management on construction sites. The results show that the accuracy of identifying scaffold units reached 76%, and the successful identification of cross braces using Hough Transform achieved a rate of 85%. Such identification can effectively issue warnings when cross braces are not installed or have been removed. By automating safety inspections, safety personnel can focus on handling more complex safety issues and prevent accidents caused by human error.

## Acknowledgment

The authors thank the National Science and Technology Council, Taiwan, for supporting this research through research grant Nos. MOST 109-2622-E-002-027, MOST 110-2622-E-002-039 and NSTC 111-2622-E-002-041.